\documentclass[12pt,a4paper]{article}
\usepackage{setspace}
\usepackage{inputenc}
\usepackage{amsmath}
\usepackage{amsfonts}
\usepackage{amssymb}

\usepackage[numbers]{natbib}

\usepackage{graphicx}
\usepackage[margin=3cm]{geometry}
\usepackage{pdflscape}
\usepackage{rotating}
\usepackage[english]{babel}
\usepackage{lipsum}
\usepackage{float}
\usepackage{sectsty}
\partfont{\large}
\sectionfont{\large}
\subsectionfont{\small}
\usepackage{parskip}
\usepackage{hyperref}
\hypersetup{hidelinks}
\usepackage{array}
\usepackage{authblk}
\usepackage [autostyle, english = american]{csquotes}
\MakeOuterQuote{"}
\usepackage[toc,page]{appendix}
\usepackage{longtable}
\usepackage{ntheorem}
\theoremseparator{:}

\usepackage{pdfpages}
\usepackage{subcaption}

\title{Deep literature reviews: an application of fine-tuned language models to migration research}

\author[a]{Stefano M. Iacus}
\affil[a]{Harvard University, USA}
\author[b,c]{Haodong Qi\footnote{Correspondence: haodong.qi@mau.se}}
\author[b]{Jiyoung Han}
\affil[b]{Malmö University, Sweden}
\affil[c]{Stockholm University Demography Unit, Sweden}
\date{}

\begin{document}

\maketitle

\begin{abstract}
    This paper presents a hybrid framework for literature reviews that augments traditional bibliometric methods with large language models (LLMs). By fine-tuning open-source LLMs, our approach enables scalable extraction of qualitative insights from large volumes of research, enhancing both the breadth and depth of knowledge synthesis. To improve annotation efficiency and consistency, we introduce an error-focused validation process in which LLMs generate initial labels and human reviewers correct misclassifications. Applying this framework to over 20,000 scientific articles on human migration, we demonstrate that a domain-adapted LLM can serve as a “specialist” model -- capable of accurately selecting relevant studies, detecting emerging trends, and identifying critical research gaps. Notably, the LLM-assisted review reveals a growing scholarly interest in climate-induced migration. However, existing literature disproportionately centers on a narrow set of environmental hazards (e.g., floods, droughts, sea-level rise, and land degradation), while overlooking others that more directly affect human health and well-being, such as air and water pollution or infectious diseases. This imbalance highlights the need for more comprehensive research that goes beyond physical environmental changes to examine their ecological and societal consequences, particularly in shaping migration as an adaptive response. Overall, our proposed framework demonstrates the potential of fine-tuned LLMs to conduct more efficient, consistent, and insightful literature reviews across disciplines, ultimately accelerating knowledge synthesis and scientific discovery. \\


\end{abstract}

\section*{Introduction}
Reviewing the state-of-the-art knowledge is a cornerstone of scientific research, essential for mapping frontiers, identifying gaps, and shaping research agendas. Traditionally, literature reviews have been labor intensive, requiring significant time and effort to analyze vast amounts of scholarly text. The exponential growth of academic publishing has only intensified this challenge, underscoring the need for more efficient and scalable approaches. To address this, we introduce a hybrid framework that enables researchers to conduct literature reviews which are both large-scale and in-depth. By fine-tuning large language models (LLMs), our approach integrates bibliometric analysis with qualitative literature review, enhancing both the breadth and depth of knowledge synthesis.

Bibliometric analysis provides a structured quantitative method for exploring research trends, identifying knowledge gaps, and analyzing collaboration networks by leveraging metadata such as citations, authorships, and keywords \citep{DONTHU2021285}. It has also become a valuable tool for mapping the mobility patterns of academic scholars \citep{zhao2023gender}. However, while effective for large-scale analysis, bibliometric methods fail to capture the deep semantic meaning embedded in research content. This limitation extends to the identification of relevant works, as keyword-based queries often yield incomplete or inaccurate search results. In contrast, qualitative literature reviews offer in-depth and critical analyses, facilitating a nuanced understanding of theories, concepts, and research findings. However, they are time-consuming, labor-intensive, and impractical for exhaustive reviews. In this article, we demonstrate how fine-tuning LLMs can bridge these two approaches, generating deeper insights from large volumes of academic literature.

The advent of LLMs has not only transformed the field of natural language processing (NLP) but also opened new frontiers in research and industry. LLM-based applications have already demonstrated their potential in fields such as clinical decision-making \citep{Hager2024}, financial analysis \citep{Gentzkow2019}, and legal bias detection \citep{Chalkidis2020}. However, their application in literature reviews (particularly in integrating qualitative insights with quantitative bibliometric analysis) remains underexplored. This article addresses this gap by introducing a scalable approach that leverages LLMs for deep and systematic synthesis of knowledge. By bridging computational efficiency with nuanced content analysis, our method enhances the extraction of meaningful insights from large-scale academic literature, ultimately accelerating scientific discovery.

Despite their transformative potential, LLMs present critical challenges, including algorithmic bias, transparency, and reproducibility concerns \citep{Abdurahman2024,Bisbee2024,Fields2024,Hagendorff2023}. These issues are particularly pronounced in proprietary (closed-source) models, such as those developed by OpenAI. While proprietary LLMs are often at the cutting edge of performance, their generalization capabilities may not translate optimally to specialized domains, as they are trained on general text data. To ensure greater transparency, reproducibility, and customization, this article focuses on fine-tuning open-source models rather than relying on proprietary alternatives.

Fine-tuning (FT) is the process of adapting pre-trained LLMs to specific applications using a smaller, domain-specific dataset, rather than training a model from scratch. This technique not only reduces computational costs but also enhances the model’s relevance for specialized tasks \citep{alizadeh2025open}. While fine-tuned LLMs receive less public attention and investment compared to large proprietary models, they have demonstrated capabilities that match or even surpass them in certain domains
\citep{Bucher2024,Carammia2024,MathavRaj2024FineTL}. In the context of literature reviews, fine-tuning offers significant advantages: it enables LLMs to better understand domain-specific terminology and conceptual nuances, improving their ability to retrieve relevant scientific works, classify topics, and summarize research content. Furthermore, it bridges the gap between qualitative and computational methods, allowing researchers to extract deep insights from large-scale publication databases that would otherwise be infeasible to analyze manually.

Fine-tuning LLMs requires high-quality human-annotated data for training and validation. Traditionally, researchers recruit and train human coders (e.g., research assistants) to annotate documents and generate labeled data for model training and validation. However, this approach is costly and time-consuming. Alternatively, researchers can rely on crowd-sourcing platforms such as MTurk, CrowdFlower, and Figure Eight. However, recent evidence suggested that annotation quality from crowd workers has declined \citep{Chmielewski2019} and their agreement with expert annotations can be low \citep{Fabbri2020}. Inter-coder disagreement poses another challenge in building gold-standard training datasets, often arising from rigid codebooks, coder fatigue, and cognitive biases \citep{Carammia2024}. The extent of disagreement varies, trained annotators may have a disagreement rate of over 20\%, while MTurk crowd workers can reach 44\% \citep{Gilardi2023}.

To mitigate these issues, this article introduces a novel error-focused annotation approach. Instead of asking annotators to select correct labels from predefined options, we ask them to reject incorrect predictions made by LLMs. This method leverages human sensitivity to errors over correctness, a well-documented phenomenon known as error salience \citep{Harsay2012} or cognitive and negativity biases \citep{holroyd2002neural, Ito1998, Mobbs2015} meaning that people are more attuned to spotting mistakes than confirming correctness. By focusing on identifying inaccuracies, this approach improves annotation reliability, reduces inter-coder disagreement, and enhances the overall quality of training data which is necessary for developing robust fine-tuned LLMs in literature review applications.

\section*{A framework of deep literature reviews using LLMs}
\begin{figure}
\centering
\includegraphics[width=1 \linewidth]{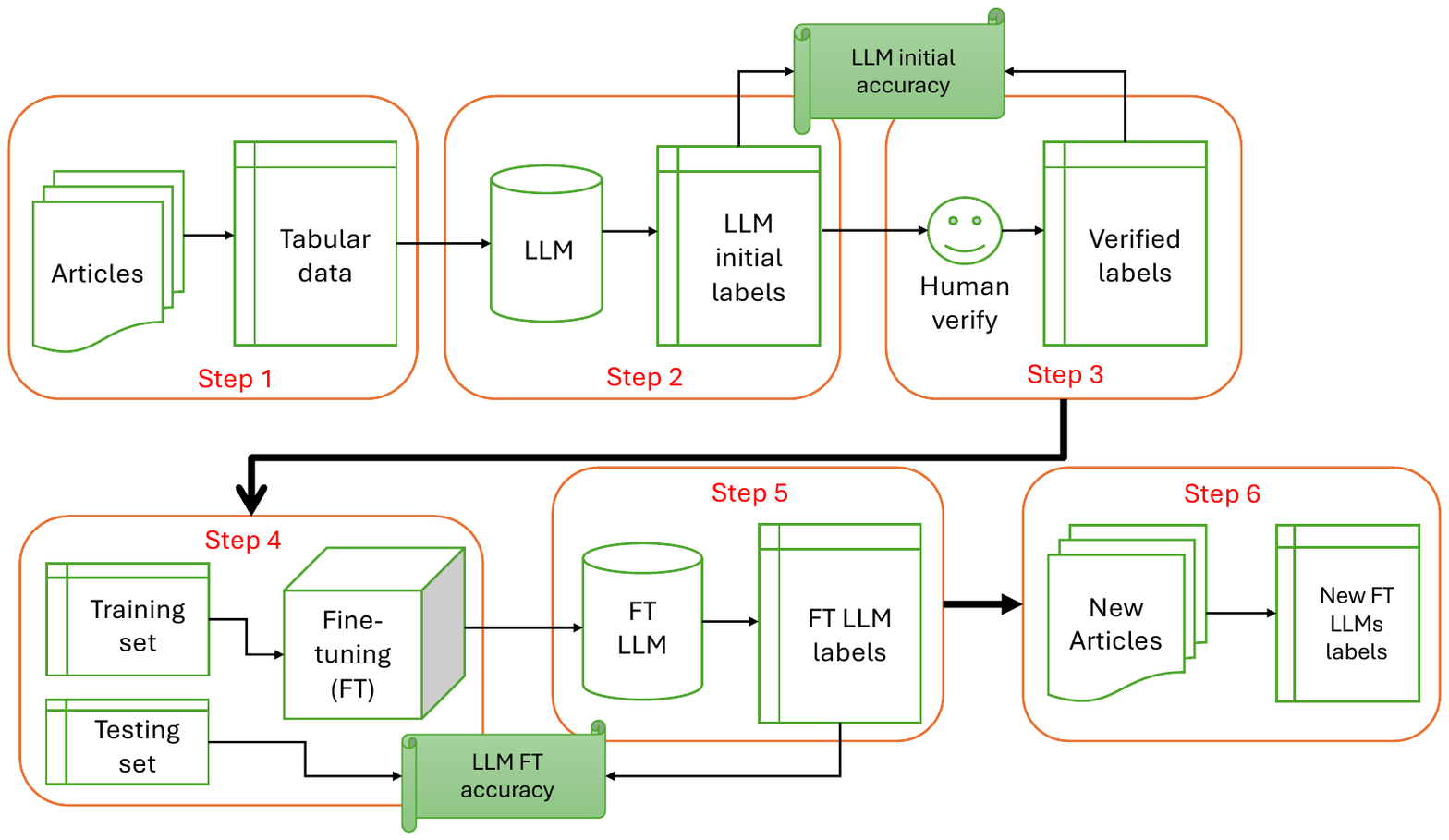}
\caption{Framework of deep literature reviews using fine-tuned LLMs}
\label{fig:methodFramework}
\end{figure}

Figure \ref{fig:methodFramework}
illustrates the workflow of LLM-assisted literature reviews. This framework can be adapted to various tasks that extract qualitative insights from scientific works, such as summarizing articles or generating unstructured responses to research-related questions. However, evaluating the quality of generated summaries or open-ended answers requires significant human effort. To alleviate this burden, our framework prioritizes structured question answering, where LLMs generate labels that best describe an article’s content. For example, in the use case presented in the next section, we instructed LLMs to classify whether an article focuses on human migration and mobility. Additionally, we asked LLMs to generate labels identifying the methodological approaches used, the migration drivers discussed, and the specific climatic or environmental hazards examined in each study. This structured approach streamlines verification, improves accuracy, and enhances the efficiency of literature analysis.

Compared to traditional bibliometric analysis, our approach offers two key advantages. First, a fine-tuned LLM classifier improves the relevance of search results by filtering articles based on specific topics more effectively. For example, in the use case below, we demonstrate that LLM-based searches can more accurately identify articles related to human migration. In contrast, traditional keyword-based queries often retrieve irrelevant results, such as studies on animal and bird migration. Second, our approach enhances the depth of literature reviews. Traditional bibliometric analysis primarily relies on structured metadata, such as authorship, cross-citations, affiliations, and countries. In contrast, our method extracts deeper insights from the unstructured textual content of research articles, enabling a more comprehensive understanding of the literature.

The proposed framework for deep literature reviews consists of six sequential steps:

Step 1: Data Processing. Literature is retrieved from scientific databases (e.g., Web of Science, Scopus) and filtered using metadata such as publication type, year, and academic field. The retrieved documents are then structured into a tabular format with key identifiers (DOI), text fields (title, abstract), and metadata (authors, affiliations, citations). A small random sample is selected for fine-tuning, while the full dataset is preserved for large-scale classification in Step 6.

Step 2: AI-Crowd Classification. Researchers define classification questions in advance based on their specific interests. For example, in the use case below, we asked: "Is an article about human migration and mobility?" and "What migration drivers does each article discuss?" Based on these predefined questions, researchers can use one or more open-source LLMs to generate initial labels (i.e., structured answers). Depending on the classification questions, these labels may be mutually exclusive (assigning only one label per article) or inclusive (allowing multiple labels per article).

Step 3: Human Verification. Human coders (experts and research assistants) review LLM-generated initial labels, rejecting misclassifications. Multiple coders may assess the same abstracts to evaluate inter-coder agreement. The verified labels are compared against AI predictions to evaluate the accuracy of initial classification.

Step 4: Fine-tuning. Verified labels in Step 3 are split into training and testing sets. The training set will be used to fine-tune open-source LLMs, enhancing their classification performance.

Step 5: Validation. The fine-tuned models classify articles in the testing set, and their outputs are validated against human-verified labels to assess performance.

Step 6: Scaling. Once validated, the fine-tuned models are deployed to classify the full dataset, as well as newly published literature, generating structured labels for further analysis.

When validating model performance, the choice of evaluation metrics depends on the number of candidate labels and whether they are mutually exclusive. For mutually exclusive classification, standard metrics such as Balanced Accuracy, Precision, Sensitivity, Specificity, and F1 Score can be applied to assess the model’s ability to correctly classify instances into a single category. In cases where an article can be assigned multiple labels, more advanced metrics are required. Commonly used measures include Hamming Loss \citep{Hamming1950}, which quantifies the proportion of incorrect label assignments, and the Jaccard Index \citep{Jaccard1901}, which evaluates the similarity between predicted and true label sets.

\section*{An application to migration and mobility studies}
Here, we demonstrate the application of our deep literature review framework in the field of human migration and mobility studies. Specifically, we retrieved 22,267 published articles from Web of Science (WoS) containing keywords semantically linked to migration and mobility. We then instruct the LLM  to perform  four classification tasks :

\begin{itemize}
    \item \textbf{Relevance Classification:} Determining whether an article is actually about human migration and mobility. This step helps assess the accuracy of keyword-based search queries in WoS, identifying potential irrelevant results.
    \item \textbf{Methodological Classification:} Identifying whether a study employs \textbf{quantitative methods} in its analysis.
    \item \textbf{Migration Drivers:} Extracting labels indicating the migration driver(s) discussed in each article.
    \item \textbf{Climate/Environmental Hazards:} Identifying which climate or environmental hazards are analyzed.
\end{itemize}

It is important to note that the labels in the first two classification tasks are mutually exclusive, we therefore apply the standard accuracy metric to evaluate the performance of LLM fine-tuning. The latter content-related questions are multi-label classifications, thereby we quantify the proportion of correct label assignments using the Jaccard Index \citep{Jaccard1901}.

For this analysis, we randomly draw 5,545 out of 22,267 retrieved articles and apply the framework depicted in Figure \ref{fig:methodFramework} to fine-tune the Llama 3.2 3B model, an open-source LLM with 3 billion parameters developed by Meta. While it is possible to manually download each article's full-text, the amount of human efforts required will be immense. To ease such a burden, we fine-tune the model based on articles' abstracts which are provided in WoS's metadata. 

\subsection*{Fine-tuning performance}
\begin{figure}
\centering
\includegraphics[width=1 \linewidth]{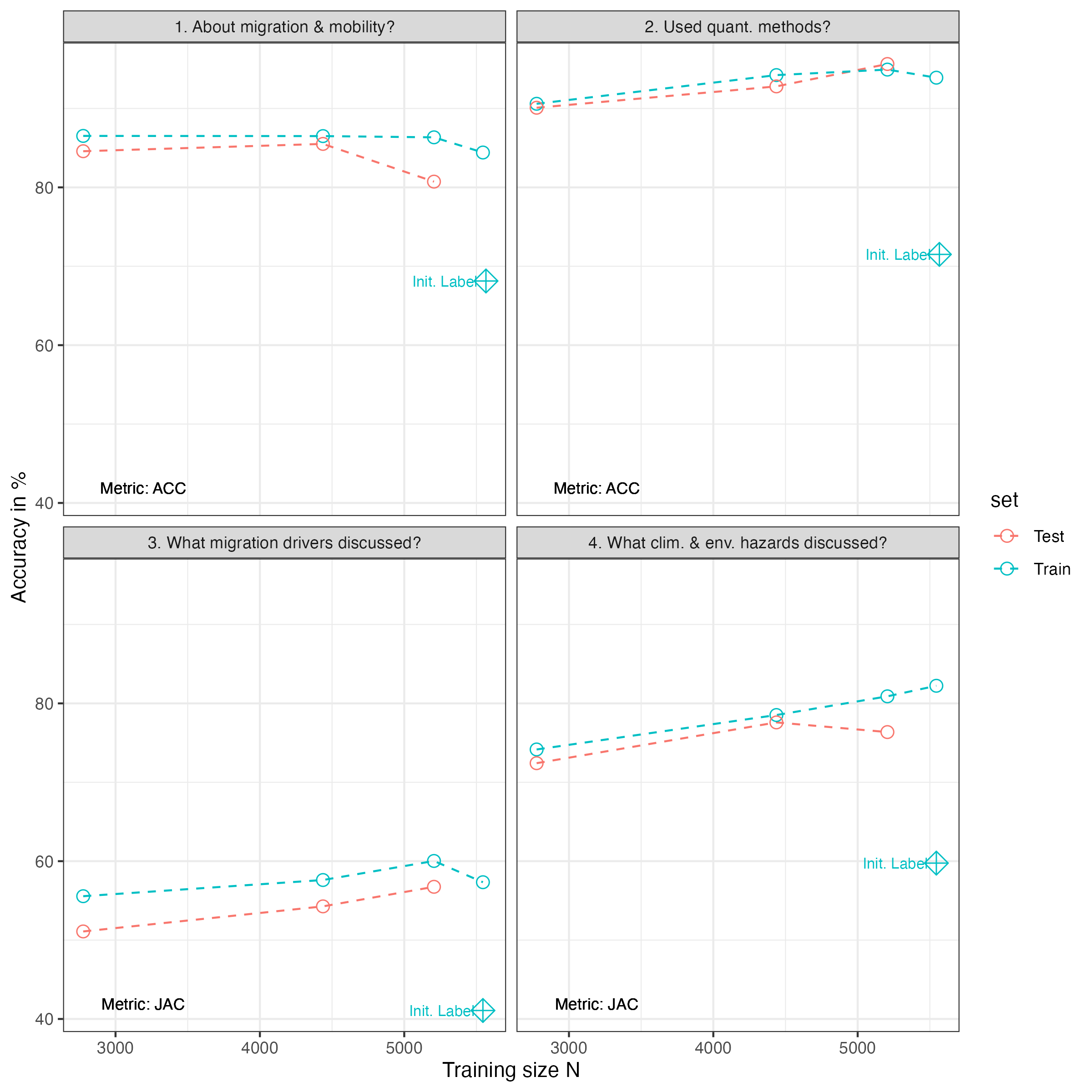}
\caption{Model performance measured by accuracy (ACC) for mutually exclusive classifications and by Jaccard Index (JAC) for multi-label classifications. Accuracy is reported for initial labels by the original Llama 3.2 3B model as well as for training and testing labels classified by the fine-tuned Llama mdoel. }
\label{fig:metric}
\end{figure}

The performance of the original Llama 3.2 3B model and its fine-tuned version is illustrated in Figure \ref{fig:metric}. The data points marked "initial label" in Figure \ref{fig:metric} correspond to Step 2 in our framework -- AI-crowding classification using the original Llama 3.2 3B model. As expected, when comparing these initial classifications with human verified labels, the accuracy is low. However, after fine-tuning, the model's performance improves significantly; as shown by the circles in Figure \ref{fig:metric}, accuracy increases by 10-25 percentage points for both training and testing datasets. 

We also examined how training sample size influences fine-tuning performance. The results suggest that binary classifications (i.e., Questions 1 and 2) achieve stable accuracy with relatively small training samples, whereas multi-label classification tasks (Questions 3 and 4) show a gradual increase in accuracy with larger training datasets. This pattern implies that the amount of training data required for effective fine-tuning depends on the complexity of the classification task.  Specifically,  binary classification tasks (e.g., identifying whether an article discusses migration or uses quantitative methods) can achieve 80\%+ accuracy with fewer than 3,000 training samples. In contrast, multi-label classification tasks (e.g., identifying migration drivers and environmental hazards) require larger datasets to achieve similar accuracy. For example, Question 4 needs over 5,000 training samples to reach 80\% accuracy, while Question 3 remains consistently low-performing, regardless of sample size.

This variation in accuracy could also be influenced by the complexity of the questions themselves. For simpler classifications, such as identifying whether an article employs quantitative methods, the model performs well because abstracts often explicitly mention statistical analysis, data sources, sampling frameworks, or empirical estimates. By contrast, for more complex classifications, such as identifying migration drivers, the model struggles. This is likely due to the inherent difficulty of understanding migration drivers -- even for human experts in the field of  migration studies; the factors driving migration are typically non-categorical, vaguely defined, and dynamic over time \citep{carammia2022forecasting,qi2023modelling, qi2023google}. As a result, many articles refer to these drivers implicitly, using terms such as cross-country wages differences, poverty, marriage, wars and conflict, making it challenging for LLMs to establish clear semantic links between text and predefined labels.

\subsection*{Scaling using the fine-tuned LLM}
As shown in the bottom-right of Figure \ref{fig:methodFramework},  once the LLM is fine-tuned, we  proceed to the final step -- scaling. In this step, we use the fine-tuned model to classify the research content of all 22,267 abstracts retrieved from WoS. In practice, this fine-tuned LLM can be applied to newly published scientific work on human migration, hence functioning as a tool for dynamic analysis of the state-of-the-art literature.

\begin{figure}
\centering
\includegraphics[width=1 \linewidth]{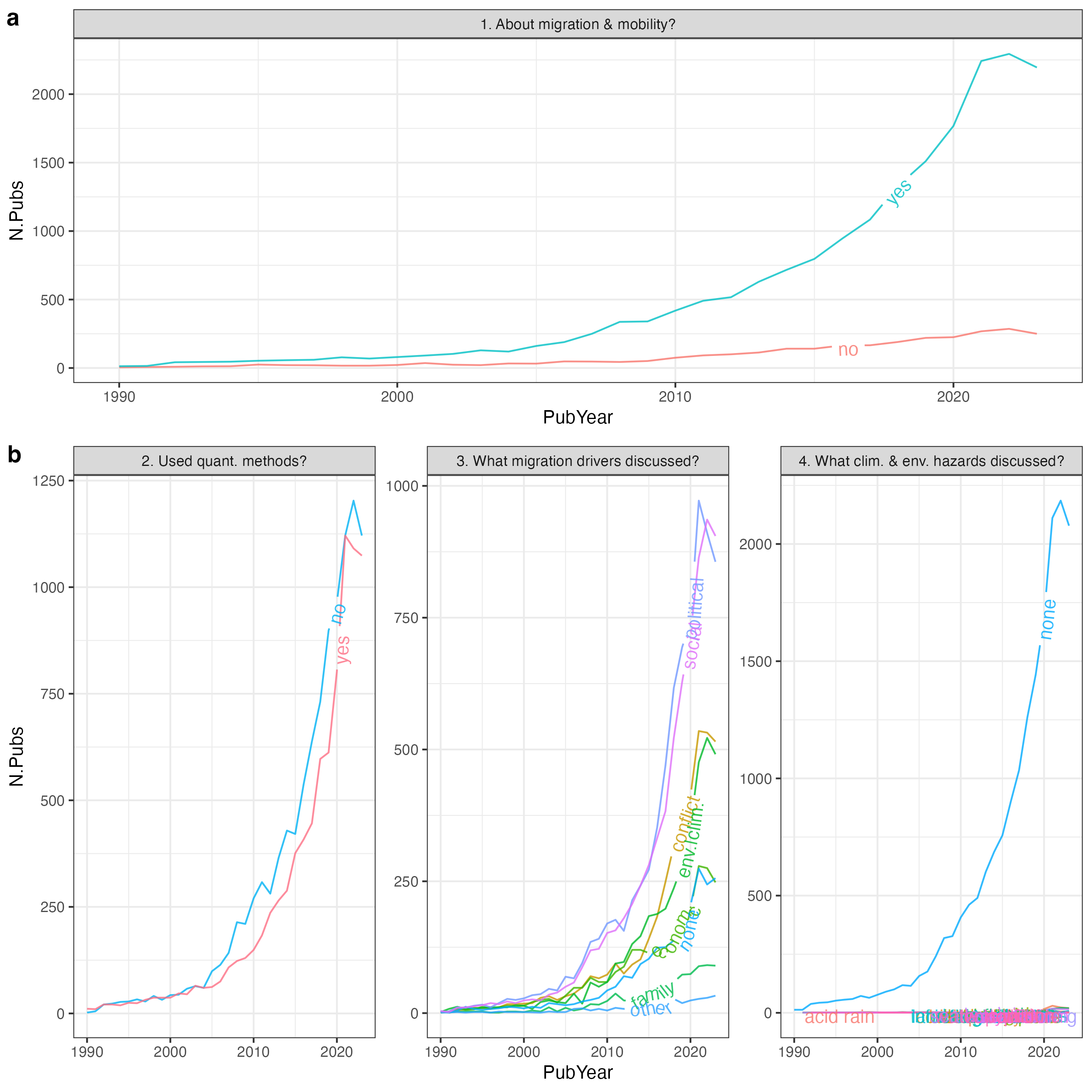}
\caption{Trends of human migration and mobility studies. Note: the data underlying panel b is a subset of 22,267 articles conditional on articles that is about human migration and mobility.}
\label{fig:trends}
\end{figure}

To demonstrate the effectiveness of our approach, we instructed the fine-tuned LLM to classify each of the 22,267 articles in terms of whether they were truly about human migration and mobility. As discussed earlier, traditional bibliographic databases such as WoS and Scopus rely on keyword-based search functions, which do not account for research context.  Consequently, of the 22,267 articles retrieved using migration-related keywords, 2,963 articles (13\%) were actually about migratory patterns of animals, birds, and genes rather than human migration. These irrelevant articles are labeled "no" in Figure \ref{fig:trends} (panel a). After filtering out non-human migration articles, we analyze the remaining 19,304 articles based on three key dimensions:  methodological approaches (qunatitative vs. qualitative), migration drivers, and climatic/environmental hazards. The results are present in Figure \ref{fig:trends} (panel b). 

The left side of panel b illustrates the number of studies that applied qualitative vs. quantitative methods. While both types of research have grown exponentially overtime, there is a noticeable lag in the expansion of quantitative migration studies. This delay is unsurprising, as macro- and micro-level data on international migration flows and migrant populations have historically been scarce, making it challenging for researchers to conduct robust quantitative analyses. Moreover, while advances in big data and computational methods have transformed fields like biology and physics, the emergence of data-driven social science has been much slower \citep{lazer2009computational}. This slow progress is even more pronounced in migration research, where quantitative studies only began to catch up with qualitative research around 2020.

The middle section of panel b illustrates the prevalence of different migration drivers in the literature, including socio-political factors, conflict, environmental/climate factors, economic drivers, family-related reasons, and other factors. It is important to note that, since Question 3 is a multi-label classification problem, an article can be counted more than once if it discussed multiple drivers. The results indicate that social and political factors have been the focal points of  migration research. This is not unexpected because cross-border migration is a highly politicized issue with major societal and policy implications for both origin and destination countries. 

Although conflict is closely linked to social and political conditions, its role in migration has been less frequently studied in earlier literature. However, since the mid-2010s, research on conflict-induced migration has grown rapidly. This surge in interest was likely driven by the Syrian civil war and the resulting "refugee crisis" in Europe (2015-2016). 

The volume of studies examining environmental and climatic impacts on migration has also increased exponentially. This trend aligns with global discourse on climate change and human displacement. As early as 1990, the Intergovernmental Panel on Climate Change (IPCC) warned that climate change could have a major impact on migration. Since then, scholarly interest in climate migrants (or "climate refugees") has steadily increased. However, despite this growing interest, most studies do not explicitly mention specific environmental hazards. Instead, climate change is often discussed as a broad, composite factor shaping human mobility. This pattern is evident in the right side of panel b, where the majority of studies did not specify particular climatic hazards.

\begin{figure}
\centering
\includegraphics[width=1 \linewidth]{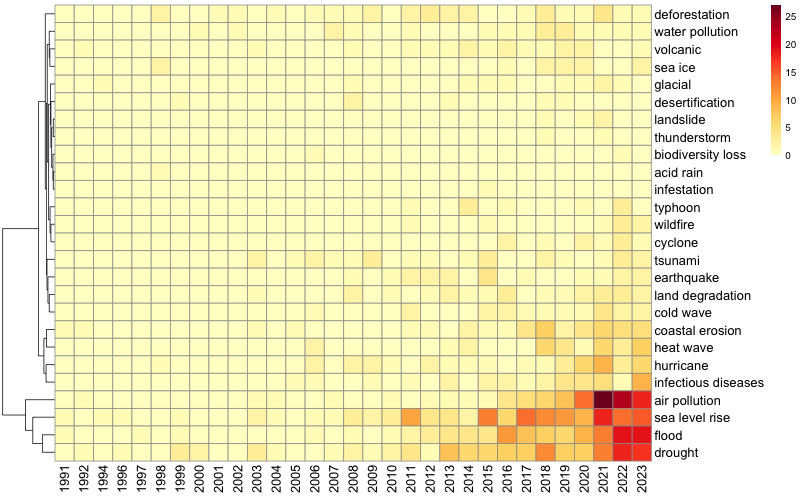}
\caption{Climatic/environmental hazards discussed in migration research. Note: the data underlying this figure is a subset of 22,267 articles conditional on i) the  articles are about human migration and mobility; and ii) the articles addressed climate/environment as a migration driver.}
\label{fig:hazHeatmap}
\end{figure}

To gain a clearer picture of which environmental hazards have been explicitly discussed in migration literature, we examined a subset of 1,350 articles that specifically addressed climate- or environmental-related drivers. Using Ward.D2 clustering, we identified the most commonly discussed hazards: air pollution, sea level rise, flood, and drought (see Figure \ref{fig:hazHeatmap}). Moreover, since late-2010s, increasing attention has also been given to costal erosion, heat wave, hurricane, infectious diseases. 

\begin{figure}
\centering
\includegraphics[width=1 \linewidth]{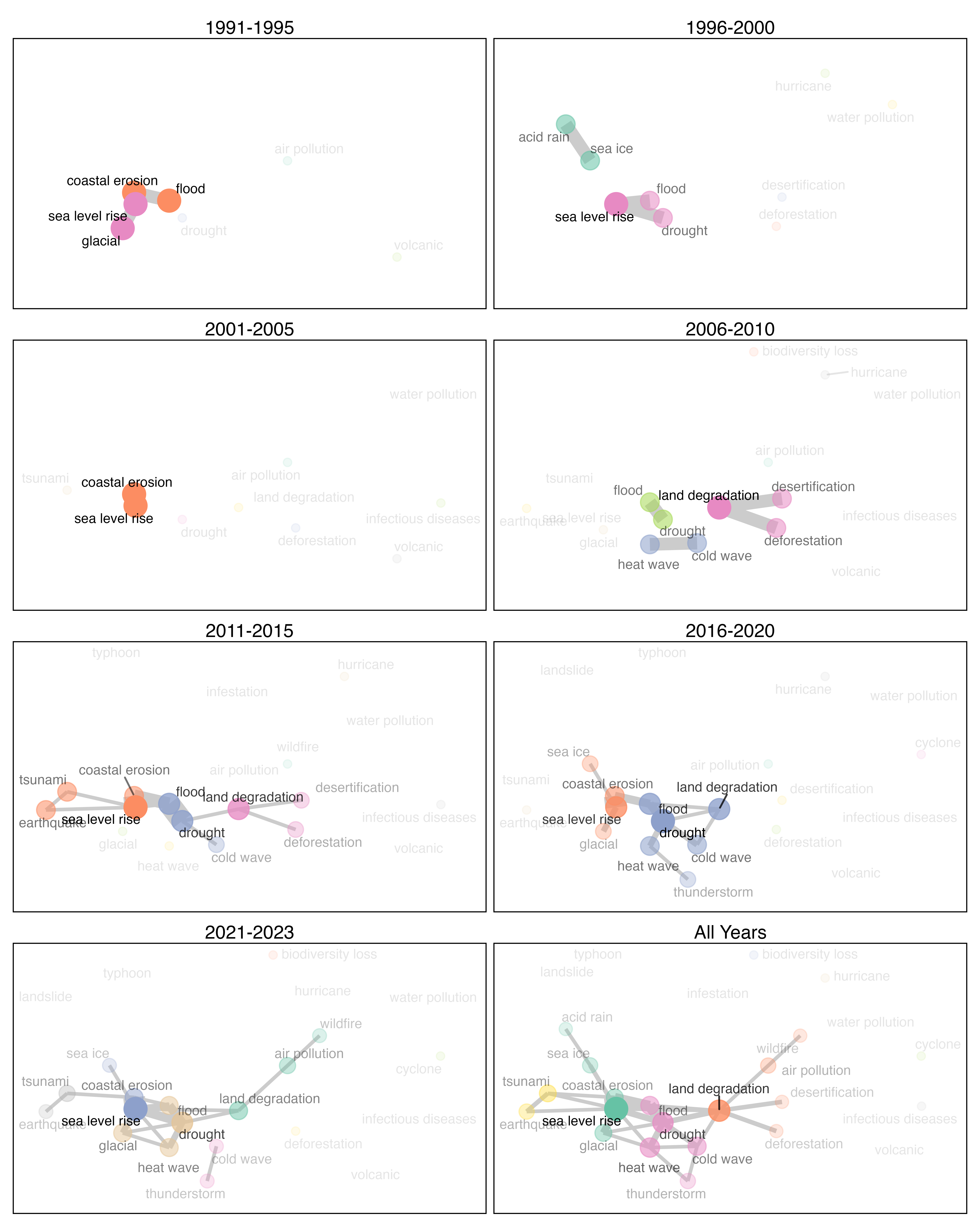}
\caption{The degree to which climatic/environmental hazards discussed jointly in a given article. Note: the data underlying this figure is a subset of 22,267 articles conditional on i) the  articles are about human migration and mobility; and ii) the articles addressed climate/environment as a migration driver.}
\label{fig:HazardNetwork}
\end{figure}

To further investigate how different hazards are jointly discussed in the migration literature, we conducted a network analysis. Figure \ref{fig:HazardNetwork} illustrates the extent to which various hazards co-occur within a given article. Each color represents a cluster of hazards that are more likely to appear together. Node size indicates centrality, which measures the relative structural importance of each hazard across the network \citep{Hevey01012018}; in other words, it reflects how strongly a hazard is connected to others. The width of the edges denotes the degree of co-occurrence between pairs of hazards.

It is evident that the network is dynamic and expands over time. An interesting pattern emerges: the network initially comprises a small set of nodes predominantly focused on hydro-climatic hazards but gradually evolves through linkages with a broader array of environmental factors. Particularly in the 1990s and early 2000s, articles tended to concentrate on wet conditions such as flooding, sea-level rise, and coastal erosion. Since 2006, the connection between hydro- and dry-climatic hazards -- especially between floods and droughts -- has become more prominent. This linkage has facilitated the network’s expansion to include additional environmental concerns such as land degradation, desertification, deforestation, and air pollution.

As shown in the final panel of Figure \ref{fig:HazardNetwork}, the existing migration literature (published during 1991-2023) increasingly addressed a wide range of interlinked environmental and climatic drivers. Notably, floods and droughts were frequently examined together and often associated with land degradation and sea-level rise. However, despite this expansion, certain hazards -- such as biodiversity loss, water and air pollution, and infectious diseases -- that may directly affect human health and well-being remain weakly connected or entirely absent. This gap underscores the need for further research that moves beyond physical environmental changes to examine their broader ecological and human consequences, particularly in terms of their impact on migration intentions and adaptive responses.

\section*{Summary and Discussion}
This article introduced a hybrid framework for conducting deep literature reviews by leveraging open-source large language models (LLMs). Our approach enhances traditional bibliometric analysis by incorporating qualitative insights extracted from research content, thereby expanding both the breadth and depth of knowledge synthesis. Unlike proprietary commercial models such as ChatGPT, our framework provides greater transparency and reproducibility through the fine-tuning of open-source LLMs, facilitating open scientific inquiry and ensuring greater control over model behavior.

Additionally, this study introduced a novel approach to combine LLMs with human oversight, overcoming key challenges in traditional annotation tasks. Manual annotation is often labor-intensive, time-consuming, and prone to inter-coder disagreement due to cognitive biases, rigid codebooks, and coder fatigue. These issues are particularly pronounced in high-dimensional classification tasks, such as categorizing environmental hazards into 40+ distinct types. Our framework mitigates these challenges by employing LLMs for initial classification, allowing human reviewers to focus on rejecting misclassifications rather than selecting from predefined categories. This error-focused annotation approach capitalizes on human sensitivity to errors rather than correctness, significantly reducing cognitive load and improving both efficiency and accuracy.

Using migration research as a case study, we demonstrated that a small, open-source LLM can function as "specialist" model after fine-tuning with domain-specific datasets . These fine-tuned models outperform traditional keyword-based search methods (such as those used in Web of Science and Scopus) in identifying relevant academic works. Beyond this enhanced search capabilities, our framework enables structured content-based analysis, providing deeper insights into the landscape of migration research.

Our large-scale review of over 20,000 scientific works revealed a significant gap in migration studies: the limited exploration of climatic and environmental hazards. Despite an exponential increase in research on environmental drivers of migration (see the panel b in Figure \ref{fig:trends}), the field has yet to systematically examine how different types of hazards influence migration. As illustrated in Figure \ref{fig:hazHeatmap}, existing research is disproportionately focused on a narrow set of hazards (such as air pollution, floods, droughts, and sea-level rise), while the human impacts of other critical risks, including desertification, biodiversity loss, landslides, and land degradation, remain largely unexplored.

It is crucial to recognize that migration responses can vary significantly depending on the type of hazard considered \citep{hoffmann2020meta}. These effects can be both positive and negative, potentially balancing out the overall impact of climate change. However, when conclusions are drawn from research that overemphasizes a limited set of hazards, the broader narrative becomes distorted. This selective focus can reinforce misleading claims about mass climate-induced migration, perpetuating the myth of "climate migration" as an impending security crisis \citep{boas2019climate}. 

To address  this research gap, a more comprehensive and interdisciplinary approach is required. First, data infrastructure must be improved allowing migration scholars to move beyond commonly studied hazards like floods and droughts. This requires high-resolution environmental and migration datasets capable of capturing a wider range of hazards and their long-term effects on human mobility. Second, stronger collaboration among migration researchers, climate scientists, and data analysts is essential for developing integrated models that account for the complexity and heterogeneity of migration responses. Finally, despite its deep roots in the social sciences, migration research should embrace more cutting-edge methodologies such as machine learning, deep learning, and remote sensing to improve the analyses of environmental stressors and their impacts on migraiton. By broadening the scope of analysis and fostering interdisciplinary partnerships, the field can develop a more comprehensive understanding of how migration responds to environmental hazards.

To conclude, this article introduced a novel framework that integrates fine-tuned LLMs with traditional bibliometric analysis, enhancing the efficiency and depth of literature reviews. While we demonstrated its effectiveness in identifying emerging trends and critical gaps in migration research, this approach is highly adaptable and can be applied across various research domains. By enabling scholars to conduct more efficient and insightful literature reviews, we hope this framework will accelerate scientific discovery and foster greater collaboration across disciplines.

\section*{Materials and methods}
This study utilized data from the Clarivate Web of Science (WoS) database. Our initial search query incorporated a combination of keywords related to human migration and mobility, including *human migrat*, *human mobilit*, *refugee*, *asylum*, *people*, *population*, *migrat*, *mobilit*, *displac*, *resettle*, *reloc*, *immigra*, *emigra*, *migrant*, *refugee*. We further applied the following selection criteria: document type (articles only), publication year (by 2023), and language (English). This yielded a total of 22,267 articles, of which 5,545 articles were randomly selected for fine-tuning. While manually downloading each article's full text is possible, the effort required would be immense. To alleviate this burden, we fine-tune the model using article abstracts, which are readily available in WoS metadata.

Following the workflow shown in Figure \ref{fig:methodFramework}, we first used the Llama 3.2 3B model (an open-source LLM with 3 billion parameters developed by Meta) to generate initial labels for four questions: 
\begin{itemize}
    \item Is the article about human migration and mobility?
    \item Did the article apply any quantitative methods in its analysis?
    \item What migration drivers did the article discuss? (Candidate labels include climate, environmental, conflict, economic, family, political, social, and other).
    \item What climatic/environmental hazards did the article discuss? (Using 41 standardized hazard classifications from the United Nations Office for Disaster Risk Reduction.).
\end{itemize}

To improve classification accuracy, the authors reviewed and corrected LLM-generated labels by rejecting misclassifications. This verified dataset was then split into training and testing sets at various ratios and used to fine-tune the Llama 3.2 3B model.

To assess model performance, we used two evaluation metrics. For mutually exclusive classification problems (Question 1 and 2), we used accuracy (ACC):
\begin{equation}
    ACC=\dfrac{1}{N} \sum_{i=1}^{N} \sum_{j=1}^{M} I(y_{ij}=\hat y_{ij})
\end{equation}
where, $N$ is the total number of articles and $M$ is the number of candidate labels per classification question. $y_{ij}$ is the true label which equals to 1 if article $i$ belongs to class $j$ and otherwise $0$. $\hat y_{ij}$ is the LLM-generated label. $I(.)$ is an indicator function which equals to $1$ if $y_{ij}=\hat y_{ij}$ and otherwise $0$.

For multi-label classifications (Question 3 and 4), we used Jaccard Index (JAC),
\begin{equation}
    JAC=\dfrac{1}{N} \sum_{i=1}^{N} \dfrac{\hat y_{i*} \cap y_{i*}}{\hat y_{i*} \cup y_{i*}}
\end{equation}
where, $y_{i*}=y_{i1}, y_{i2}, ..., y_{iM}$ is a $M$ length true vector containing the human-verified label(s) for article $i$. $\hat y_{i*}=y_{i1}, y_{i2}, ..., y_{iL}$ is a $L$ length predicted vector containing LLM-generated label(s).

Since the intersection is always smaller than or equal to the union, the Jaccard Index falls within the range $[0,1]$. A higher JAC indicates greater classification accuracy, with $JAC = 1 $ denoting perfect agreement and $JAC = 0$ indicating no overlap between true and predicted labels.

After fine-tuning, we used the Llama 3.2 3B model to infer research content for all 22,267 articles retrieved from WoS. Descriptive analyses were then conducted to examine patterns and trends in the generated labels. 

\bibliographystyle{ieeetr}

\begin{thebibliography}{10}

\bibitem{DONTHU2021285}
N.~Donthu, S.~Kumar, D.~Mukherjee, N.~Pandey, and W.~M. Lim, ``How to conduct a
  bibliometric analysis: An overview and guidelines,'' {\em Journal of Business
  Research}, vol.~133, pp.~285--296, 2021.

\bibitem{zhao2023gender}
X.~Zhao, A.~Akbaritabar, R.~Kashyap, and E.~Zagheni, ``A gender perspective on
  the global migration of scholars,'' {\em Proceedings of the National Academy
  of Sciences}, vol.~120, no.~10, p.~e2214664120, 2023.

\bibitem{Hager2024}
P.~Hager, F.~Jungmann, R.~Holland, K.~Bhagat, I.~Hubrecht, M.~Knauer,
  J.~Vielhauer, M.~Makowski, R.~Braren, G.~Kaissis, and D.~Rueckert,
  ``Evaluation and mitigation of the limitations of large language models in
  clinical decision-making,'' {\em Nature Medicine}, vol.~30, pp.~2613--2622,
  July 2024.

\bibitem{Gentzkow2019}
M.~Gentzkow, B.~Kelly, and M.~Taddy, ``{Text as Data},'' {\em Journal of
  Economic Literature}, vol.~57, pp.~535--574, Sept. 2019.

\bibitem{Chalkidis2020}
I.~Chalkidis, M.~Fergadiotis, P.~Malakasiotis, N.~Aletras, and
  I.~Androutsopoulos, ``{LEGAL-BERT: The Muppets straight out of Law School},''
  Oct. 2020.

\bibitem{Abdurahman2024}
S.~Abdurahman, M.~Atari, F.~Karimi-Malekabadi, M.~J. Xue, J.~Trager, P.~S.
  Park, P.~Golazizian, A.~Omrani, and M.~Dehghani, ``Perils and opportunities
  in using large language models in psychological research,'' {\em PNAS Nexus},
  vol.~3, June 2024.

\bibitem{Bisbee2024}
J.~Bisbee, J.~D. Clinton, C.~Dorff, B.~Kenkel, and J.~M. Larson, ``{Synthetic
  Replacements for Human Survey Data? The Perils of Large Language Models},''
  {\em Political Analysis}, vol.~32, pp.~401--416, May 2024.

\bibitem{Fields2024}
J.~Fields, K.~Chovanec, and P.~Madiraju, ``{A Survey of Text Classification
  With Transformers: How Wide? How Large? How Long? How Accurate? How
  Expensive? How Safe?},'' {\em IEEE Access}, vol.~12, pp.~6518--6531, 2024.

\bibitem{Hagendorff2023}
T.~Hagendorff, S.~Fabi, and M.~Kosinski, ``Human-like intuitive behavior and
  reasoning biases emerged in large language models but disappeared in
  chatgpt,'' {\em Nature Computational Science}, vol.~3, pp.~833--838, Oct.
  2023.

\bibitem{alizadeh2025open}
M.~Alizadeh, M.~Kubli, Z.~Samei, S.~Dehghani, M.~Zahedivafa, J.~D. Bermeo,
  M.~Korobeynikova, and F.~Gilardi, ``Open-source llms for text annotation: a
  practical guide for model setting and fine-tuning,'' {\em Journal of
  Computational Social Science}, vol.~8, no.~1, pp.~1--25, 2025.

\bibitem{Bucher2024}
M.~J.~J. Bucher and M.~Martini, ``{Fine-Tuned 'Small' LLMs (Still)
  Significantly Outperform Zero-Shot Generative AI Models in Text
  Classification},'' 2024.

\bibitem{Carammia2024}
M.~Carammia, S.~M. Iacus, and G.~Porro, ``{Rethinking Scale: The Efficacy of
  Fine-Tuned Open-Source LLMs in Large-Scale Reproducible Social Science
  Research},'' 2024.

\bibitem{MathavRaj2024FineTL}
J.~MathavRaj, V.~Kushala, H.~Warrier, and Y.~Gupta, ``{Fine Tuning LLM for
  Enterprise: Practical Guidelines and Recommendations},'' {\em ArXiv},
  vol.~abs/2404.10779, 2024.

\bibitem{Chmielewski2019}
M.~Chmielewski and S.~C. Kucker, ``An mturk crisis? shifts in data quality and
  the impact on study results,'' {\em {Social Psychological and Personality
  Science}}, vol.~11, pp.~464--473, Oct. 2019.

\bibitem{Fabbri2020}
A.~R. Fabbri, W.~Kryściński, B.~McCann, C.~Xiong, R.~Socher, and D.~Radev,
  ``{SummEval: Re-evaluating Summarization Evaluation},'' July 2020.

\bibitem{Gilardi2023}
F.~Gilardi, M.~Alizadeh, and M.~Kubli, ``{ChatGPT outperforms crowd workers for
  text-annotation tasks},'' {\em Proceedings of the National Academy of
  Sciences}, vol.~120, July 2023.

\bibitem{Harsay2012}
H.~A. Harsay, M.~Spaan, J.~G. Wijnen, and K.~R. Ridderinkhof, ``{Error
  Awareness and Salience Processing in the Oddball Task: Shared Neural
  Mechanisms},'' {\em Frontiers in Human Neuroscience}, vol.~6, 2012.

\bibitem{holroyd2002neural}
C.~B. Holroyd and M.~G. Coles, ``The neural basis of human error processing:
  reinforcement learning, dopamine, and the error-related negativity.,'' {\em
  Psychological review}, vol.~109, no.~4, p.~679, 2002.

\bibitem{Ito1998}
T.~A. Ito, J.~T. Larsen, N.~K. Smith, and J.~T. Cacioppo, ``Negative
  information weighs more heavily on the brain: The negativity bias in
  evaluative categorizations.,'' {\em Journal of Personality and Social
  Psychology}, vol.~75, no.~4, pp.~887--900, 1998.

\bibitem{Mobbs2015}
D.~Mobbs, C.~C. Hagan, T.~Dalgleish, B.~Silston, and C.~Pravost, ``The ecology
  of human fear: survival optimization and the nervous system,'' {\em Frontiers
  in Neuroscience}, vol.~9, Mar. 2015.

\bibitem{Hamming1950}
R.~W. Hamming, ``{Error Detecting and Error Correcting Codes},'' {\em Bell
  System Technical Journal}, vol.~29, pp.~147--160, Apr. 1950.

\bibitem{Jaccard1901}
{Jaccard, Paul}, ``Étude comparative de la distribution florale dans une
  portion des alpes et du jura,'' 1901.

\bibitem{carammia2022forecasting}
M.~Carammia, S.~M. Iacus, and T.~Wilkin, ``Forecasting asylum-related migration
  flows with machine learning and data at scale,'' {\em Scientific Reports},
  vol.~12, no.~1, p.~1457, 2022.

\bibitem{qi2023modelling}
H.~Qi and T.~Bircan, ``Modelling and predicting forced migration,'' {\em Plos
  one}, vol.~18, no.~4, p.~e0284416, 2023.

\bibitem{qi2023google}
H.~Qi and T.~Bircan, ``Can google trends predict asylum-seekers’ destination
  choices?,'' {\em EPJ Data Science}, vol.~12, no.~1, p.~41, 2023.

\bibitem{lazer2009computational}
D.~Lazer, A.~Pentland, L.~Adamic, S.~Aral, A.-L. Barab{\'a}si, D.~Brewer,
  N.~Christakis, N.~Contractor, J.~Fowler, M.~Gutmann, {\em et~al.},
  ``Computational social science,'' {\em Science}, vol.~323, no.~5915,
  pp.~721--723, 2009.

\bibitem{Hevey01012018}
D.~H. and, ``Network analysis: a brief overview and tutorial,'' {\em Health
  Psychology and Behavioral Medicine}, vol.~6, no.~1, pp.~301--328, 2018.
\newblock PMID: 34040834.

\bibitem{hoffmann2020meta}
R.~Hoffmann, A.~Dimitrova, R.~Muttarak, J.~Crespo~Cuaresma, and J.~Peisker, ``A
  meta-analysis of country-level studies on environmental change and
  migration,'' {\em Nature Climate Change}, vol.~10, no.~10, pp.~904--912,
  2020.

\bibitem{boas2019climate}
I.~Boas, C.~Farbotko, H.~Adams, H.~Sterly, S.~Bush, K.~Van~der Geest,
  H.~Wiegel, H.~Ashraf, A.~Baldwin, G.~Bettini, {\em et~al.}, ``Climate
  migration myths,'' {\em Nature Climate Change}, vol.~9, no.~12, pp.~901--903,
  2019.

\end{thebibliography}

\end{document}